\documentclass[11pt,a4paper]{article}
\usepackage[dvipsnames, svgnames, x11names]{xcolor}  % 一般放得靠前

\usepackage[hyperref]{acl2021}
\usepackage{times}
\usepackage{latexsym}

\usepackage{todonotes}

\usepackage{microtype}

\usepackage{algorithmicx}  
\usepackage{algpseudocode}  
\usepackage{amsmath}  
\usepackage{color,xcolor}
\usepackage{algorithm} %format of the algorithm 

\usepackage{pifont}
\usepackage{bm}
\usepackage{amsfonts}

\usepackage{graphicx}
\usepackage[normalem]{ulem}
\usepackage{multirow}
\usepackage{amsmath}
\usepackage{url}

\usepackage{tikz}
\usepackage{subfigure}
\usepackage{xcolor}
\usepackage{tcolorbox}
\usepackage{diagbox}

\usepackage{amssymb}
\usepackage{amsopn}
\usepackage{graphicx}
\usepackage{multirow}
\usepackage{makecell}
\usepackage[english]{babel}
\usepackage{bm}
\usepackage{array}
\usepackage{setspace}
\usepackage{CJKutf8}

\aclfinalcopy % Uncomment this line for the final submission
\title{On Compositional Generalization of Neural Machine Translation}

\author{
 Yafu Li$^{\spadesuit \heartsuit}$\hspace{0.5mm}, 
 Yongjing Yin$^{\spadesuit \heartsuit}$\hspace{0.5mm}, 
 Yulong Chen$^{\spadesuit \heartsuit}$\hspace{0.5mm}, 
 Yue Zhang$^{\heartsuit \diamondsuit}$\hspace{0.2mm}\hspace{1.5mm} \\
 $^\spadesuit$ Zhejiang University\\
 $^\heartsuit$ School of Engineering, Westlake University\\
 $^\diamondsuit$ Institute of Advanced Technology, Westlake Institute for Advanced Study\\ 
 \textit{yafuly@gmail.com} \quad\textit{yinyongjing@westlake.edu.cn}\\
 \textit{yulongchen1010@gmail.com} \quad\textit{yue.zhang@wias.org.cn} \\
}
\date{}

\begin{document}
\maketitle
\begin{abstract}
Modern neural machine translation (NMT) models have achieved competitive performance in standard benchmarks such as WMT.
However, there still exist significant issues such as robustness, domain generalization, etc.
In this paper, we study NMT models from the perspective of compositional generalization by building a benchmark dataset, CoGnition, consisting of 216k clean and consistent sentence pairs. 
We quantitatively analyze effects of various factors using compound translation error rate, then demonstrate that the NMT model fails badly on compositional generalization, although it performs remarkably well under traditional metrics.
% This indicates that our dataset can serve as a testbed for methods tailored to handling compositionality in NMT.

% Moreover, we release a test set in which there are 10,800 constructed sentences containing novel compounds. 
% there are no publicly large-scale translation corpora designed for systematic compositionality study, and thus previous work has resorted to proof-of-concept experiments.
% while exhibit limited compositional generalization. 
% which is a crucial concept in language understanding.
% Compositionality i and recives incresing interests recently.
% Noisy or non-standard input text can cause disastrous mistranslations in most modern Machine Translation (MT) systems, and there has been growing research interest in creating noise-robust MT systems. 

\end{abstract}

\section{Introduction}

Neural machine translation (NMT) has shown competitive performance on benchmark datasets such as IWSLT and WMT \cite{VaswaniSPUJGKP17,Edunov:emnlp18,Liu:corr200807772}, and even achieves parity with professional human translation under certain evaluation settings \cite{DBLP:journals/corr/abs-1803-05567}. 
% Despite its recent success, NMT still faces various challenges, most notably in out-of-domain and under low resource conditions \cite{KoehnK17}. 
However, the performance can be relatively low in out-of-domain and low-resource conditions. 
In addition, NMT systems show poor robustness and vulnerability to input perturbations \cite{BelinkovB18, ChengJM19}. 
One example is shown in Table \ref{translation_engine}, where simple substitution of a word yields translation with completely different semantics.
Many of these issues origin from the fact that NMT models are trained end-to-end over large parallel data, where new test sentences can be sparse. 
% This can manifest as challenges such as domain mismatch, data inefficiency and inability to translate long sentences \cite{KoehnK17}.
% This can manifest as various errors such as under-translation, over-translation and mistranslation, severely undermining stability and reliability of NMT systems. 
\begin{table}
\centering
\small
% \setlength{\tabcolsep}{2.5pt}
% \begin{spacing}{1.0}
\begin{CJK*}{UTF8}{gbsn}
\begin{tabular}{c|c}
Input &	Translation \\
\hline
\hline
\multirow{2}{*}{Taylor \textbf{breaks} his promise} & 泰勒{\color{red}{信守}}诺言 \\
& (Taylor {\color{red}{keeps}} his promise) \\
\hline
\multirow{2}{*}{James \textbf{breaks} his promise} & 詹姆斯{\color{blue}{违反}}诺言 \\
& (James {\color{blue}{breaks}} his promise) \\
\hline
\end{tabular}
% \end{spacing}
\caption{
\label{translation_engine}
Translation samples obtained from one popular web translation engine on January 19, 2021.
}
% \footnote{https://translate.google.cn/}
\end{CJK*}
\end{table}

% , which maybe a new test sentence can be sparse.
% in an entirely data-driven fashion without imposing explicit rules of language. 
% The major problem lies in data sparsity. 

% With the development of language, new words and their combinations emerge continuously, 
% meaning there always exist samples that are never encountered during training. 
% From this perspective, any parallel corpus suffers from data sparsity in some degree. 
% Therefore, inefficiency in exploiting data limits NMT model's capability under low-resource and out-of-domain settings, and memorization without comprehending language rules prevents models from generalizing to input perturbations. 
\begin{table*}[t!]
\centering
% \scriptsize
\small
% \footnotesize
% \setlength{\tabcolsep}{2.5pt}
% \begin{spacing}{1.0}
\begin{CJK*}{UTF8}{gbsn}
\begin{tabular}{l|l|l|l}
\hline
\textbf{Dataset} & \textbf{Type} & \textbf{Source} & \textbf{Target} \\
\hline
\hline
\multirow{2}{*}{SCAN} & {\it Atoms} & jump, twice & \small  \multirow{2}{*}{ JUMP JUMP} \\
& {\it Compounds} & jump twice  \\
\hline
\multirow{3}{*}{CFQ} & {\it Atoms} & Who [predicate] [entity], directed,  Elysium & \scriptsize SELECT DISTINCT ?x0 WHERE \{ \\
& {\it Compounds} & Who directed Elysium & \scriptsize ?x0 a ns:people.person . \\
& & & \scriptsize ?x0 ns:film.director.film m.0gwm\_wy\} \\
%   SELECT DISTINCT ?x0 WHERE ?x0 a ns:people.person . \\
% & & ?x0 ns:film.director.film m.0gwm\_wy \} \\
% CFQ &  Who directed & \tiny SELECT DISTINCT ?x0 WHERE \\
% & Elysium & \scriptsize \{ ?x0 a ns:people.person . \\
% & & \scriptsize ?x0 ns:film.director.film \} \\
% & & \scriptsize m.0gwm\_wy \} \\
\hline
\multirow{3}{*}{CoGnition} & {\it Atoms} & the, doctor, he liked & \multirow{3}{*}{他喜欢的医生病了} \\
& {\it Compounds} & the doctor he liked & \\
& {\it Sentences} & The doctor he liked was sick  & \\
% ROC- & The doctor he & 他喜欢的医生 \\
% Parallel & liked was sick & 病了 \\
\hline
\end{tabular}
% \end{spacing}
\caption{
\label{scan_cfq}
Examples of SCAN, CFQ, and our CoGnition datasets.
}
\end{CJK*}
\end{table*}

Disregarding out-of-vocabulary words, a main cause of sparsity is semantic composition:
% Translation can be viewed as a process of semantic transformation, which is required to understand semantics of components and generalize their combinations. Infinite combinations are the root of data sparsity: 
given a limited vocabulary, the number of possible compositions grows exponentially with respect to the composite length.
% Although the meaning of word components brings with long-tailed words and out-of-vocabulary (OOV), in this work we focus on analyzing model's ability of generalizing various combinations. 
The ability to understand and produce a potentially infinite number of novel combinations of known components, namely \textit{compositional generalization} \cite{chomsky1957,montague1974d,Lake:icml18,Keysers:iclr2020}, 
has been demonstrated deficient in many machine learning (ML) methods \cite{JohnsonHMFZG17,Lake:icml18,BastingsBWCK18,LoulaBL18,Russin19}.
% \citet{Keysers:iclr2020} characterize compositional generalization with two basic elements: \textit{atom}, i.e., primitive elements in train set, and \textit{compound}, i.e., novel compositions of \textit{atoms}.
% Regarding compositional generalization, two aspects are involved:

In this paper, we study compositional generalization in the context of machine translation. 
For example, if ``{\it red cars}'' and ``{\it blue balls}'' are seen in training, a competent algorithm is expected to translate ``{\it red balls}'' correctly, even if the phrase has not been seen in training data. Intuitively, the challenge increases as the composite length grows.
Recently, several studies have taken steps towards this specific problem. They either use a few dedicated samples (i.e., 8 test sentences) for evaluation \cite{Lake:icml18,LiZWH19,ChenLYSZ20}, or make simple modifications in sampled source sentences such as removing or adding adverbs, and concatenating two sentences \cite{Raunak,fadaee-monz-2020-unreasonable}.
Such experimental data is limited in size, scope and specificity, and the forms of composition are coarse-grained and non-systematic. 
As a result, no qualitative conclusions have been drawn on the prevalence and characteristics of this problem in modern NMT.
% On a fine-grained level, it still remains uncertain what we have achieved overall and what fundamental
% changes each milestone technique has brought.

% Differing from previous work \cite{Fadaee:acl2020, Raunak} which are limited in simple experiment settings, we construct a dedicated dataset and analyze in a more quantified and systematic way. 

We make a first large-scale general domain investigation, constructing the CoGnition dataset 
(\textbf{Co}mpositional \textbf{G}eneralizati\textbf{on} Mach\textbf{i}ne \textbf{T}ranslat\textbf{ion} Dataset), a clean and consistent parallel dataset in English-Chinese,
% , for English-Chinese translation. 
along with a synthetic test set to quantify and analyze the compositional generalization of NMT models. 
In particular, we define frequent syntactic constituents as \textit{compounds}, and basic semantic components in constituents as \textit{atoms}. 
% \textit{Atoms} are manually selected to compose novel \textit{compounds}, which are further embedded in sentential contexts to form test sentences. 
In addition to the standard training, validation and test sets, the CoGnition dataset contains a compositional generalization test set, which contains novel compounds in each sentence, so that both the generalization error rate can be evaluated, and its influence on BLEU \cite{PapineniRWZ02} can be quantified. 
Our compositional generalization test set consists of 2,160 novel compounds, with up to 5 atoms and 7 words.
In this way, generalization ability can be evaluated based on compound translation error rate. 

Empirical results show that the dominant Transformer \cite{VaswaniSPUJGKP17} NMT model faces challenges in translating novel compounds, despite its competitive performance under traditional evaluation metrics such as BLEU. 
In addition, we observe that various factors exert salient effects on model's ability of compositional generalization, such as compound frequency, compound length, atom co-occurrence, linguistic patterns, and context complexity. 
The CoGnition dataset along with the automatic evaluation tool are realesed on \href{https://github.com/yafuly/CoGnition}{https://github.com/yafuly/CoGnition}.
% https://github.com/yafuly/CoGnition
% We will release our dataset along with the automatic evaluation tool upon acceptance.

% NMT performance. challenges. in Figure \ref{exp} google translation.

% NMT mechanism. model conditional sequence distribution, sparsity. cause the challenges above.
% must have unseen sentences.

% if model can understand unit and how to compose, it can translate. rare words and oov not included in this paper. 
% we focus on compositional generalization.

% translation is semantic transformation. include unit understand and composition. Without rare words, composition is root of sparsity, composition boom. so model should be computational.
% although several related work xxx, we differently xxxx.

% we propose a dataset.xxxx

% experiment and analysis show that xxx.

% James breaks his promise &	詹姆斯违反了他的诺言 \\
% \hline
% James breaks his promise. &	詹姆斯违背了诺言。 \\
% \hline

\section{Related Work}

\paragraph{Analysis of NMT.}
% As the NMT models has yielded strong empirical performance, 
% Many efforts have been devoted to
Our work is related to research analyzing NMT from various perspectives.
% For example, some studies conduct different levels of 
There has been much linguistic analysis of NMT representations \cite{Shi:emnlp2016, Belinkov:ijcnlp2017, Bisazza:emnlp2018}, interpretability \cite{Ding:acl2017, He:emnlp2019, Voita:emnlp2019}, and attention weights \cite{voita-etal-2019-analyzing, Michel:nips2019}.
Robustness is also an important research direction.
Work has shown that NMT models are prone to be negatively affected by both synthetic and natural noise \cite{Belinkov:iclr2018, Cheng:acl2018, Ebrahimi:coling2018}.
For better exploration of robust NMT, \citet{Michel:emnlp2018} propose an MTNT dataset containing several types of noise. 
% By contrast, the source sentences in our dataset are clear from noise and the test set contains novel compounds that do not appear in the training set.
\citet{Wang:aclL20} provide in-depth analyses of inference miscalibration of NMT resulting from the discrepancy between training and inference. 
% Different from the above, we conduct systematic analysis of NMT models from the perspective of compositional generalization.
Our work is in line but we discuss robustness from the perspective of compositional generalization.

% There are two studies most related to us. 
In this respect, \citet{Lake:icml18} propose a simple experiment to analyze compositionality in MT, followed by \citet{ChenLYSZ20} and \citet{LiZWH19}. 
Specifically, they introduce a novel word ``\textit{dax}'', and their training data contains a single pattern of sentence pairs (e.g. ``\textit{I am daxy}'', ``\textit{je suis daxiste}'') while the test set contains different patterns. However, their work is limited in that there are only 8 sentences in the test set.
\citet{Raunak} observe a performance drop on a dataset of concatenated source sentences.
\citet{Fadaee:acl2020} modify source sentences by removing adverbs, substituting numbers, inserting words that tend to keep syntax correct (e.g. ``\textit{very}''), and changing the gender, and find unexpected changes in the translation.
% The main difference is that their modifications are extremely simple such as removing adverbs or substituting numbers while we create novel constituents.
% Different from using WMT for training, we construct a clean and consistent parallel data set to minimize effects of others factors.
In contrast to these studies, we quantitatively measure compositionality of NMT under compound translation error rate.

Translation involves various challenges such as low-frequency words, polysemy and compositional complexity. In this work, we \textbf{focus} on how the NMT model generalizes to complex compositions in a controllable setting and minimize the effects of the other factors.

\paragraph{Compositional Generalization.}
% Human language and thought are characterized by systematic compositionality, the algebraic capacity to understand and produce a potentially infinite number of novel combinations from known components \cite{Chomsky, Montague}.
Neural networks have been shown sample-inefficient, requiring large-scale training data, which suggests that they may lack compositionality \cite{Lake:icml18}.
\citet{Lake:icml18} introduce the SCAN dataset to help study compositional generalization of neural networks, which has received increasing interests \cite{Russin:corr1904, Roberto:ac2019, Li:emnlp2019, Lake:nips2019, Andreas:acl2020, Gordon:iclr2020}.
Various benchmarks have been proposed including in the area of visual reasoning \cite{Johnson:cvpr2017, Hudson:cvpr19}, mathematics \cite{Saxton:iclr19}, and semantic parsing (CFQ) \cite{Keysers:iclr2020}.
However, no benchmark has been dedicated to machine translation in practice. 
We fill this gap by introducing a dataset with 216,000 instances and an average sentence length of 9.7 tokens.

% The main difference between ours and SCAN is that samples in the SCAN dataset are just simple language commands (e.g. jump twice) while our dataset is composed of bilingual sentence pairs with rich semantic constituents.
% The CFQ dataset is automatically generated based on some pre-defined rules, and the target is not a natural language.
% language has highly different properties to a natural language.

\begin{figure*}[!t]
\centering
\includegraphics[width=1.0\linewidth]{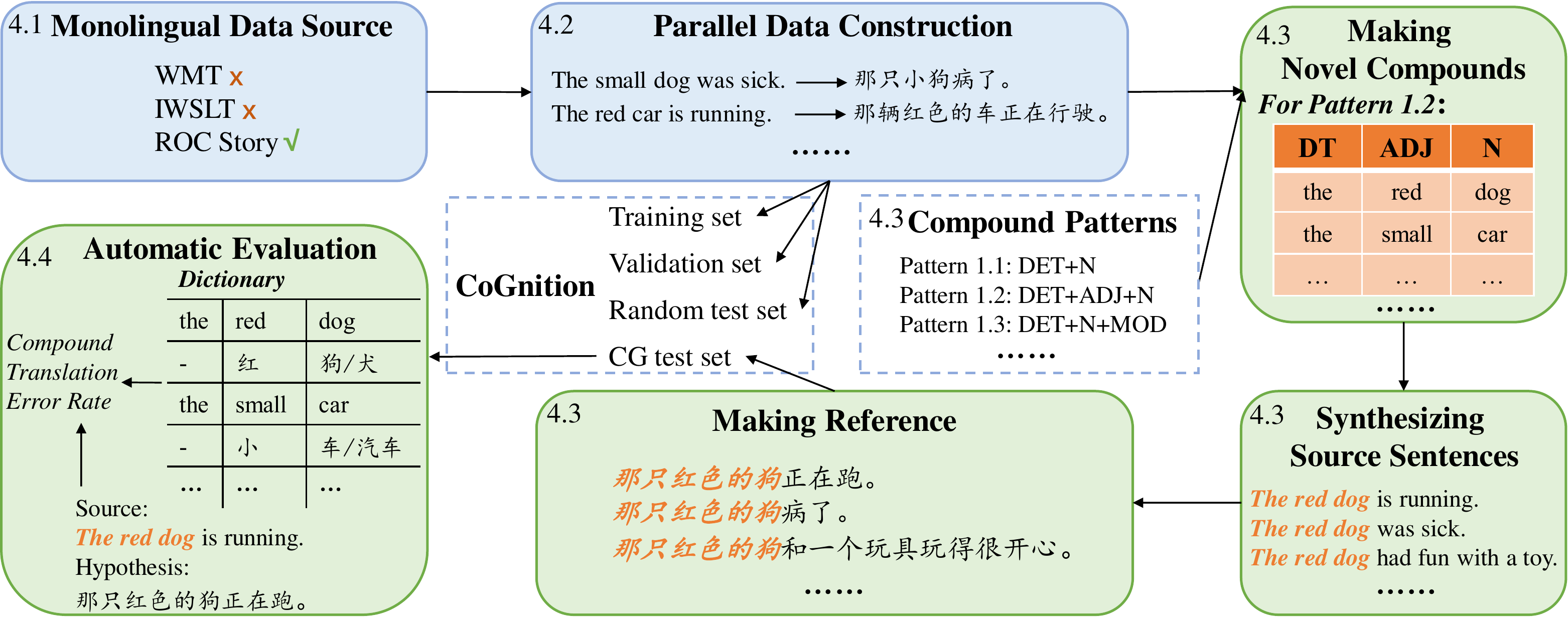}
\caption{
\label{data_flow}
Summary of dataset construction.}
\end{figure*}

\begin{CJK*}{UTF8}{gbsn}
\section{Problem Definition}\label{Problem Definition}
% \todo{need a figure}
% \todo{why not use CFQ rule}
% 1 what is compositional generalizaiton
% 2 atom and compound
% 3 from SCAN to MT

% In the context of learning from a set of training examples, 
Following \citet{Keysers:iclr2020}, {\it compositional generalization} is defined as the 
capacity to systematically generalize to novel combinations of components which are learned sufficiently during training.
Key elements to measure compositional generalization include {\it atoms} and {\it compounds}.
Specifically, \textbf{atoms} are primitive elements in the train set whereas \textbf{compounds} are obtained by composing these atoms. 
% of which distributions are similar in the train and test sets, and compounds are obtained by composing these atoms. 
The research question is whether neural models perform well on unseen compounds. 
% in the test set which are composed of the atoms of the train set.
Take Table \ref{scan_cfq} for example, in the SCAN dataset, the atoms are simple commands such as ``\textit{jump}'' and the composite command ``\textit{jump twice}'' is a compound. 
In the CFQ, the compounds are questions such as ``\textit{Who directed Elysium}'', and the atoms correspond to the primitive elements in the questions such as the predicate ``\textit{directed}'', the question patterns ``\textit{Who [predicate] [entity]}'' and the entities ``\textit{Elysium}''.

In theory, compounds in MT can be defined as phrases, sentences or even document. In practice, however, we want to control the number of atoms in a novel compound for quantitative evaluation. 
In addition, it can be highly difficult to construct a large-scale dataset where novel compounds are sentences of practical sizes (the number of synthesized sentences increases exponentially with their length) while ensuring their grammatical correctness.
Therefore, we constrain {\it compounds} to syntactic constituents, and define {\it atoms} as basic semantic components in constituents according to syntactic and semantic rules for forming constituents \cite{partee1995lexical}.
As a result, we randomly assign multiple sentential contexts for investigating each novel compound. Table \ref{scan_cfq} shows a contrast between our dataset and existing datasets for compositional generalization in semantics. 
% Back to Figure \ref{scan_cfq}, the compound in a sample of our data is a noun phrase ``\textit{The doctor he liked}'' and the atoms are its basic components. In this paper, we measure compositional generalization of NMT models by quantitatively analyzing compound and sentence translation quality.
% error rates as well as sentence translation quality.
% Although the trained NMT model has seen all atoms during training, it still translates the novel compound incorrectly. 

Mistakes caused by weakness in computational generalization can be easily found in state-of-the-art NMT models. 
% For machine translation, state-of-the-art NMT models have achieved high BLEU scores in frequently used benchmarks. 
% However, the competitive BLEU scores cover up some problems.
In particular, we train a Transformer-based model \cite{VaswaniSPUJGKP17} on WMT17 En-Zh Dataset \footnote{http://www.statmt.org/wmt17/}.
% For example, one sentence in the standard test set, ``\textit{in the dermatologist clinic, a 60-year-old man \textbf{came for consultation with her little grandson}}'', is translated into ``在皮肤科诊所, 一位60岁的男子来咨询她的小孙子'' (English: \textit{in the dermatologist clinic, a 60-year-old man \textbf{came to consult her little grandson}}). 
% Atoms in the compound ``\textit{came for consultation with her little grandson}'' are of high frequency in the training set, 862 times for ``\textit{grandson}'' and more than 90,000 for the other ones. 
One sentence in the standard test set, ``\textit{but the problem is , with the arrival of durant , thompson 's \textbf{appearance rate} will surely decline , which is bound to affect his play}'', is translated into ``但问题是, 随着杜兰特的到来, 汤普森的外表肯定会下降, 这一定会影响到他的表演'' (English: \textit{but the problem is , with the arrival of durant , thompson 's will surely \textbf{look worse} , which is bound to affect his play}). 
The novel compound ``\textit{appearance rate}'' is composed of two atoms (i.e., ``\textit{appearance}'' and ``\textit{rate}''), both with a high frequency of more than 27,000 times in the training set. 
% For example, one sentence in the standard test set, ``\textit{remember to apply sunblock half an hour in advance, because sunscreen products take time to \textbf{be completely absorbed into the skin}}'', is translated into ``记住提前半小时使用防晒霜, 因为防晒霜产品需要时间才能完全吸收皮肤'' (English: \textit{remember to apply sunblock half an hour in advance, because sunscreen products take time to \textbf{completely absorb the skin}}).
% Atoms in the compound ``\textit{be completely absorbed into the skin}'' are highly frequent in the training set, with a minimum frequency of $4098$ for the word ``\textit{skin}''. 
% However, the model never sees the whole compound during training and thus outputs incorrect translation with completely different semantics. 
However, the sentence semantics is completely distorted due to the failure of semantic composition, which is possibly influenced by the context word ``\textit{play}''.
More importantly, as the overall translation highly overlaps with the reference, the model achieves a high score in similarity-based metrics such as BLEU, demonstrating that fatal translation errors can be overlooked under traditional evaluation metrics.

% In the example above, there is a polysemous word ``{\it appearance}'', which we do not consider in our constructed dataset in order to isolate the most salient factors for investigation. The next section will show more reasons why WMT is not ideal for serving as a dataset to evaluate compositional generalization.

\end{CJK*}
% In particular, we train an NMT model on WMT17 En-Zh dataset and find a sample in the test set: the model makes severe mistakes in translating ``\textit{the 63-year-old has now been...}'' to 
% \begin{CJK*}{UTF8}{gbsn}
% ``这位63岁的儿童...'' 
% \end{CJK*}
% (English: the 63-year-old \textbf{child}).
% Except this phrase, the model makes totally correct translation, and 
% However, the model achieves a relatively high BLEU score on this sample, 
% such fatal mistakes can be submerge by model's competitive performance under commonly used BLEU metrics. 
% For example, a sample in WMT14 En-De corpus ``the 63-year-old has now been...'' gets xxx BLEU but is translated severely incorrectly to 
% makes severe mistakes in translating one test sample in WMT17 En-Zh,

% though the words ``63'', ``year'', and ``old'', all occur frequently in the training data.
% % but the composition ``63-year-old'' .
% We attribute this mistake to lack of compositional generalizaiton.

\begin{table*}
\centering
\small
% \setlength{\tabcolsep}{2.5pt}
% \begin{spacing}{1.0}
\begin{tabular}{l|c|c}
\hline
\textbf{Pattern \#}	& \textbf{Composition}	& \textbf{Example} \\
\hline
\hline
Pattern 1.1	& DET+N	& all the sudden \textbf{the waiter} screamed in pain . \\
\hline
Pattern 1.2	& DET+ADJ+N	& one day \textbf{another lazy lawyer} snapped and broke every window in the car . \\
\hline
Pattern 1.3	& DET+N+MOD	& \textbf{each doctor he liked} was talking to a friend on the phone . \\
\hline
Pattern 1.4	& DET+ADJ+N+MOD	& \textbf{every smart lawyer at the store} decided to go back next week . \\
\hline
Pattern 2.1	& V+DET+N	& she said she \textbf{liked the building} ! \\
\hline
Pattern 2.2	& V+DET+ADJ+N	& he soon \textbf{met the special girl} named taylor . \\
\hline
Pattern 2.3	& V+DET+N+MOD	& she \textbf{took the child he liked} out to enjoy the snow . \\
\hline
Pattern 2.4	& V+DET+ADJ+N+MOD	& when taylor \textbf{saw the dirty car he liked} , he was amazed . \\
\hline
Pattern 3.1	& P+DET+N	& taylor felt really awful \textbf{about the bee} . \\
\hline
Pattern 3.2	& P+DET+ADJ+N	& \textbf{inside the small apartment} were some of my old toys . \\
\hline
Pattern 3.3	& P+DET+N+MOD	& taylor forgot \textbf{about the chair on the floor} ! \\
\hline
Pattern 3.4	& P+DET+ADJ+N+MOD	& he jumped from the bench \textbf{towards the large airplane on the floor} . \\
\hline
\end{tabular}
% \end{spacing}
\caption{
\label{talbe_pattern}
Compound patterns in the CG test set.  Compounds are in bold and shown in sentence context.
}
\end{table*}

\section{Dataset}
% introduction
% We establish an English-Chinese translation dataset to quantify and analyze the compositional generalization of NMT systems.
% We construct a dataset for analysis and measure NMT's compositionality.
% Concretely, we explanate the motivation of our data source selection in Section \ref{data source}.
% In Section \ref{parallel data} we describe the construction of parallel data including adopting machine translation and post-editing.
% This is followed by a detailed description of how to create the test set in Section \ref{test set}. 

Figure \ref{data_flow} gives an overview of our data construction process.
We first source monolingual data (Section \ref{data source}), and then build parallel data based by translation (Section \ref{parallel data}). 
Then we synthesize a test set of novel compounds (Section \ref{test set}), and offer an automatic evaluation method (Section \ref{auto eval}).

\subsection{Monolingual Data Source}\label{data source}
Our goal is to focus on compositional generalization and minimize the influence of additional factors such as polysemy~\citep{berard2019machine}, misalignment~\citep{munteanu2005improving}, and stylistic problems~\citep{hovy2020can}.
The dataset should ideally have following characteristics.
First, the vocabulary size should be small and contain only words of high-frequency in order to avoid problems caused by rare words. In other words, variety of composition should come from combining different frequent words instead of word diversity, as suggested in \cite{Keysers:iclr2020}.
Metaphorical words, which can increase the translation difficulty, should be excluded.
Second, source sentences should not be too long or have complex syntactic structures. As a result, a sentence can be translated literally, directly, and without rhetoric.
Third, the corpus size should be large enough for training an NMT model sufficiently.

Widely-adopted corpora such as parallel data released on WMT and IWSLT\footnote{https://wit3.fbk.eu/} have large vocabularies and also contain noisy sentences and rich morphology \cite{WMT19niu}, which do not fully meet our goal. 
We choose Story Cloze Test and ROCStories Corpora~\citep{mostafazadeh-etal-2016-corpus, mostafazadeh2017lsdsem} as our data source. 
The dataset is created for commonsense story understanding and generation, and consists of $101903$ \textit{5-sentence} stories.
These stories are rather simple in items of vocabulary and syntax, but still contain rich phrases.
In addition, the topic is constrained to daily life.

Since the vocabulary size of $42,458$ is large, we select the top $2,000$ frequent words as our vocabulary and extract sentences where the words are exclusively from the restricted vocab. 
Moreover, sentences that are longer than 20 words are removed. 
In this way, we finally obtain $216,246$ sentences for parallel data construction.
More detailed statistics including comparison to WMT and IWSLT data are shown in Appendix \ref{Data Statistics}.
% characteristics make the data source a potential testbed for compositionality research. 

% why select ROC instead of WMT or IWSLT.
% 直接翻译。一次多义，对齐缺失。去除这些因素，主要研究组合。
% 数据简单，翻译知识简单
% 中短句。ＲＯＣ本身对自己的描述。

\subsection{Parallel Data Construction}\label{parallel data}
% NMT systems tend to generate more deterministic and consistent translation than human translators \cite{Kim:emnlp2016, Gu:iclr2018, Zhou:iclr2020}, which can serve as a basis for building a translation corpus with high statistical consistency.
% Hence, 
We take an MT post-editing method to construct parallel data, first using a public translation engine to obtain model-generated translations, 
% translate collected source data from English to Chinese, 
and then requesting expert translators to post-edit them.
% the model-generated translations. 
The following aspects are highlighted:
% % To ensure the quality of automatic machine translations, we randomly sample $10,000$ pairs from this model-generated corpus, and manually evaluate them. We find that over $97.5\%$ translated sentences are fluent, grammatical, semantic persistent, and the left $2.5\%$ sentences are only partially not correct.
% MT + human post edit, why? manual translation 1. costly 2. not consistent, paraphrase. MT system output is more deterministic cite(NAT).
%%%%%%%%%
% 1.只看comp是否翻译对，不在乎整句话。分类问题。测试模型对某个cp翻译的准确度。最重要的标准。
%%%%%%%Yulong 这是对模型的人工评测, 不是做数据集，不写这里。
% Given our dataset is limited to scale and vocabulary size, we do not want to introduce out-of-vocabulary words. Therefore, we ask annotators to pay extra attention to :
\begin{itemize}
    \item Ensure the fluency of translations.
    \item Ensure word-level matching between translated sentences and source sentences. Typically, every word should be correctly translated, without omission for legibility.
\end{itemize}

Finally, we obtain a parallel dataset of $216,246$ sentences in \textbf{CoGnition}, and randomly split it into three subsets: $196,246$ sentence pairs for {\bf training}, $10,000$ sentence pairs for {\bf validation}, and $10,000$ sentence pairs as the {\bf random test set}.
In addition to the above split, we additionally make a \textbf{compositional generalization test set}, which is described in the next section.
% Table X summarizes the statistics of the final data.(???)
% the training, validation, and test set contain $196,246$, $10,000$, and $10,000$ sentence pairs, respectively. .
% The lexical and syntactic knowledge in this corpus is more simple that can be easily learned by a NMT model.(???)
    % \item Translators are asked to only use knowledge inside the training set,  and paraphrase words from a dictionary that we collect from the training data.
    % \item Translators are asked to translate the sentence literally instead of freely.
    % \item Translators are asked to translate the meaning of every word in the original text as much as possible, and not to omit words for legibility (such as omitting pronouns).
% To verify our assumption, we trained a Transformer model on this corpus, and tested its performance.
% As shown in Table XX, Transformer has achieved a satisfying performance ($70\%$ \textsc{bleu-1} on test set).
% It suggests that Transformer fits perfectly on this corpus.
% With the other characteristics mentioned, we believe it can be a suitable corpus for research on compositional generalization.

% 处理，抽取一个子集。
% 训练测试，分割。７０ｂｌｅｕ。模型适应
% mt后校验，ａｔｏｍ的翻译尽可能一致。
% 在这么一个模型可以适应的很好的数据集，探究组合繁华。
% 50句多万句中，只取了由前2000最高词频的词组成的句子。
% 最大句长不超过20.。
% 得到21w，分train-19, dev-1,test-1。
% ７０ｂｌｅｕ。模型在测试集翻译没有问题了。（模型拟合非常好）

\begin{table*}[!t]
\centering
\small
% \setlength{\tabcolsep}{2.5pt}
% \begin{spacing}{1.0}
\begin{tabular}{c|c}
\hline
\textbf{Type} & \textbf{Candidates} \\
\hline
\hline
DET	& the, every, any, another, each \\
\hline
\multirow{2}{*}{N} & car, dog, girl, doctor, boyfriend, apartment, child, sandwich  \\
& chair, farm, building, hat, waiter, airplane, lawyer, peanut, farmer, clown, bee \\
\hline
ADJ	& small, large, red, special, quiet, empty, dirty, lazy, smart, fake, silly \\
\hline
% \hline
MOD	&  he liked, at the store, on the floor\\
\hline
\multirow{2}{*}{V} & took, told, found, asked, saw, left, gave, lost, liked  \\
& woke, stopped, invited, met, caught, heard, hated, watched, visited, chose \\
\hline
\multirow{2}{*}{P} & to, for, on, with, from, about, before, like, around  \\
& inside, without, behind, under, near, towards, except, toward \\
\hline
\end{tabular}
% \end{spacing}
\caption{
\label{atom_candidates}
Atoms used in constructing compounds, sorted by frequency in the training set.
}
\end{table*}

\subsection{Compositional Generalization Test Set}\label{test set}
% To investigate model performance in respect to compositional generalization, we further set up a synthetic test set consisting of synthesized source sentences and their translations.
% In addition to the random test set, 
We manually construct a special test set dedicated for evaluation of compositional generalization, by synthesizing new source sentences based on novel compounds and known contexts.

% sending them to expert translators for manual translations.

% simulated

% \paragraph{Definition of Atom and Compound}
% % 句法分析，np, vp, pp三个成分最多。
% % 根据这三个成分，来构造句子，
% % 1. NP. 
% % 2. VP. 及物动词 + NP
% % 3. PP. 介词 + PP

% % ~\footnote{https://github.com/nikitakit/self-attentive-parser}
% Instantiation of Compound and Atom
% \paragraph{Methodology}

% To determine which constituents to, we use Berkeley Parser \cite{Kitaev-2018-SelfAttentive} to parse source sentences and count the constituents frequency.
\paragraph{Designing Compound Patterns}
We use Berkeley Parser to obtain constituent trees \cite{Kitaev-2018-SelfAttentive}. In CoGnition, noun phrases (NP), verb phrases (VP) and positional phrases (PP) are three most frequent constituents, accounting for $85.1\%$ of all constituents,
% In addition, NP and VP are necessary to form a sentence \cite{yule2020study}.
% form the simplest possible sentences. 
and thus we construct compounds based on them. 
% As for selection of atoms, 
According to syntactic and semantic rules \cite{partee1995lexical}, we choose basic semantic components as our atoms including determiners (DET), nouns (N), verbs (V), prepositions (P), adjectives (ADJ), and postpositive modifiers (MOD).
Specifically, postpositive modifiers include prepositional phrases and relative clauses, and can contain multiple words. 
We consider them as a single atom due to their semantic inseparability. 
In this way, we generate 4 compound patterns for NP, VP, and PP, respectively, which are listed in Table \ref{talbe_pattern} with corresponding examples.
% Besides NP and VP, prepositional phrases (PP) are much more frequent than other constituents in our dataset. Therefore, we construct compounds based on these three types of constituents.
% which three are chosen for compound construction.

% In phrase structure grammar \cite{chomsky2002syntactic}, constitutions often reveal hierarchical structures. 
% Concretely, constituents are organized in a hierarchical structure, by embedding inside one another to form larger constituents \cite{carnie2012syntax}. As introduced in \cite{brinton2010linguistic}, most VP contain NP and every PP contain NP. 
% Therefore we start from creating NP compounds and then further extend to VP and PP compounds.
% Hence we take NP as an example to explain the definition of {\it atom} and how atoms construct compounds.
% Next, we define units in a constituent as {\it atom} according to semantic compositionality theory \cite{}. , , 

%  the basic units of NP include . 
% Besides, modifiers like , can be further added to enrich semantic meaning of NP. 
% Given that REL and PP often follow after N, we collectively refer to these two as postpositive modifiers (MOD). When constructing VP and PP compounds, we append an extra atom of verbs (V) or prepositions (P) to NP compound accordingly. Finally, all these atoms, i.e., DET, N, ADJ, MOD, V and P, are used in compound construction.

% practice
% \paragraph{Composition of Novel Compounds}
\paragraph{Making Novel Compounds}
We use Stanza \cite{qi2020stanza} to obtain POS tagging for each word in training sentences. 
We construct novel compounds by first selecting atom candidates with relatively consistent translation in the training set. 
The frequency of candidate atoms covers a wide range from $34$ to $73518$. We list full set of atom candidates in Table \ref{atom_candidates}. 
For constructing compounds, we enumerate all possible combinations of atoms according to the patterns in Table \ref{talbe_pattern}, 
and then remove those that are ungrammatical or likely to cause ethic issues, obtaining 2,160 compounds finally. 
We do not deliberately make all compounds unseen, yet only $0.93\%$ of them appear in the training data.

% \todo{details of composition}
% For the atoms in each pattern, we select several word segments of different frequencies from the training set. 
% More importantly, the selected word segments should have consistent translation, and specific 
% and no ambiguity.
% We sample 5, 5, 8, 2 word segments for DET, ADJ, NN, MOD training data.
% And we use those words to fill their corresponding atom of a noun phrase compound (for a noun phrase, adjective and modifier is not necessary, their slots can be none), 
% the maximum number of new NP compounds is $720$. 
% selection is presented in Appendix \ref{}. 
% \todo{most of compounds are novel}

\paragraph{Synthesizing  Source Sentences}
We embed the compounds in specific context to form complete source sentences. 
Concretely, we first apply Berkeley Parser on the training sentences to obtain sentence templates, where certain constituents are replaced by placeholders according to their constituent types, e.g., ``\textit{NP-placeholder spent a lot of time to set up a wedding .}''. 
Then we select 5 sentence templates for each constructed compound accordingly, 
so that every compound can be evaluated under 5 different contexts. 
To distinguish from VP and PP, we put NP compounds only in sentences with the placeholder outside VP and PP.
% serving as subjective.

% For each new compound, we first randomly choose $20$ source sentences from the training data. Then, for each sentence, we replace the segment of which the constituents is identical to that of the compound.
% When processing the compounds belonging to NP, we substitute the NP that is not inside VP and PP.
% which often contain NP.
% This ensures the removed noun phrase is not a part of verb or prepositional phrase.
% Then we insert the diagnostic noun phrase compound into those contexts, and thus create $5$ diagnostic sentences.

\paragraph{Making Reference} To maintain statistical consistency, target translations of synthetic sentences are also obtained using the same MT post-edit approach. 
% We organize expert translators to generate gold translation for the synthetic sentences. 
In addition to the annotation principles listed in \ref{parallel data}, we set several additional rules:
\begin{itemize}
    \item Filter sentences with ethical issues and replace them with other synthetic ones.
    % \item Translators are asked to watch for ethical issues. Sentences like  should be excluded in the final data set.
    \item Ensure the accuracy of compound translation.
\end{itemize}

Finally, we obtain a compositional generalization test set (\textbf{CG test set}) of $10,800$ parallel sentences.
The final dataset statistics is shown in table \ref{cognition statistics}.

\begin{table}
\centering
\small
% \setlength{\tabcolsep}{2.5pt}
% \begin{spacing}{1.0}
\begin{tabular}{c|c}
\hline
\textbf{Split} & \textbf{\# Samples}  \\
\hline
\hline
Training set & 196,246  \\
\hline
Validation set & 10,000  \\
\hline
Random test set & 10,000  \\
\hline
CG test set & 10,800  \\
\hline
\end{tabular}
% \end{spacing}
\caption{
\label{cognition statistics}
Statistics of CoGnition Dataset.
}
\end{table}

\subsection{Automatic Evaluation}\label{auto eval}
% We organize expert translators to evaluate the correctness of compound translation in order to obtain detailed and reliable conclusions. 
We mainly adopt human evaluation for the experiments of this paper (Section \ref{experiments}) for ensuring reliability of findings. 
Despite its accuracy, human evaluation can be expensive.
% and not convenient for future research. 
To facilitate fast evaluation in future research, 
% To address this issue,
we introduce an automatic evaluation approach to quantify a model's generalization ability on our CG test set.
% , but it is inconvenient for research community to evaluate their proposed models. 
% More importantly, human evaluation is not consistent and comparable among different results. 
% Therefore, we present a possible way for automatic evaluation to quantify a model's generalization ability on our generalization test set.
% Considering that the whole data set will be released for future research, we propose an automatic evaluation tool to approximately measure translation accuracy of new methods on the novel compounds.

In particular, we manually construct a dictionary for all the atoms based on the training set (See Appendix \ref{Lexicon}). 
% A part of the dictionary is provided in . 
The prerequisite of correctly translating one compound is that all of the atom translations should be contained. Besides, in most cases the translation of nouns should be placed after that of other atoms. 
Based on this, we design a heuristic algorithm to determine whether compounds are translated correctly. 
With the human annotation as ground truth, our automatic evaluation tool achieves a precision of $94.80\%$ and a recall of $87.05\%$, demonstrating it can serve as an approximate alternative to human evaluation.

% we can approximately calculate the accuracy of compound translation.
% , which contains all golden target words. 
% If a translation contain all words in the corresponding dictionary, the compound in the source sentence is regarded as correct translation.

\section{Experiments}
We conduct experiments on CoGnition dataset and perform human evaluation on the model results.
\label{experiments}
% We train an NMT model on Roc-Parallel dataset and evaluate it on both the random test set and our manual test set. 
% We manually evaluate the presence and characteristics of compositional generalization using CoGnition dataset.
% To quantitatively measure of the model's capability of compositional generalization，
% we report BLEU scores and accuracy  compound translation, as a .
\subsection{Settings}
We tokenize the English side using Moses tokenizer and do not apply byte pair encoding (BPE) \cite{Sennrichbpe} due to the small vocabulary (i.e., 2000).
The Chinese sentences are segmented by jieba segmenter\footnote{https://github.com/fxsjy/jieba}. 
We employ BPE with 3,000 merge operations, generating a vocabulary of 5,500 subwords. 

We focus on Transformer \cite{VaswaniSPUJGKP17} because of its state-of-the-art performance on machine translation \cite{DBLP:conf/emnlp/EdunovOAG18, DBLP:journals/corr/abs-2104-06022, DBLP:journals/jmlr/RaffelSRLNMZLL20, DBLP:conf/iclr/ZhuXWHQZLL20, DBLP:journals/corr/abs-2008-07772} and better performance on existing compositional generalization dataset \cite{daniel-etal-2019-towards}. We implement our model using BASE configuration provided by Fairseq \cite{ott-etal-2019-fairseq}. The model consists of a 6-layer encoder and a 6-layer decoder with the hidden size 512. 
We tie input and output embeddings on the target side. 
The model parameters are optimized by Adam \cite{adam}, with $\beta_{1} = 0.1$, $\beta_{2} = 0.98$ and $\epsilon = 10^{-9}$. 
% We follow the same learning strategy suggested by \citet{VaswaniSPUJGKP17}. 
The model is trained for 100,000 steps and we choose the best checkpoint on validation set for evaluation.

We report character-level BLEU scores using SacreBLEU \cite{sacrebleu} to measure the overall translation performance.
% report BLEU scores on each test set
% One of the mostly used metrics for machine translation is BLEU score \cite{PapineniRWZ02}. Therefore we use SacreBLEU \cite{sacrebleu} to evaluate overall performance of model translation. We 
% Another metric for measuring model's ability of compositional generalization is error rate of compound translation. 
% To this end, 
% to quantify our model's capability of generalizing to unseen compounds, 
In addition, we request expert translators to annotate the correctness of compound translation.
% During annotation of the constructed test set, 
% do not need to evaluate the whole translation and 
Translators are asked to only focus on examining whether the compound itself is translated correctly or not, disregarding errors in context. 
Specifically, a compound is correct only if its translation contains semantic meaning of all atoms and is fluent in human language.
Since each of the 2,160 compounds is provided with 5 contexts, we can compute the translation error-rate for each compound.
% by dividing the total number of contexts by the number of contexts in which compounds are not translated correctly. 
% The compound translation error-rate can be directly calculated by dividing the total number of sentences by the number of sentences of which compound is not translated correctly. 
% Since each of the 720 compounds is provided with 5 contexts, we compute compound translation error-rate 
% Besides, we compute the {\it compound-wise error rate}, i.e. average error rate for each compound under 5 contexts. 
%

% whereas compound-level accuracy is the ratio of compounds that are translated correctly in all contexts. We present both error rates in main results. We use sentence-level error rate for further analysis because it is statistically unbiased and can better quantify how well the model generalizes one compound.
% the error rate is computed in two different ways:

% During annotation of the constructed test set, annotators did not need to evaluate the whole translation and only focused on examining whether the compound itself in one sentence is translated correctly or not.
% Although the context of each sentence has been seen during training, novel compounds have salient negative effects on the whole sentence translation.

\subsection{Main Results}
\begin{CJK*}{UTF8}{gbsn}
Table \ref{tab_err_rate} shows the results. 
Besides the \textit{CG test set}, we also list results on three of its subsets, which only contain NP, VP or PP compounds respectively.
The model achieves a 69.58 BLEU score on the \textit{random test set}, which partly indicates distributional consistency and quality of the dataset. 
In comparison, the performance on the \textit{CG test set} drops dramatically by more than 20 BLEU points. 
Given that the only difference between synthetic sentences and training sentences is the unseen compounds (i.e., contexts are seen in training), 
the decrease of 20 BLEU points indicates that unseen compounds pose a significant challenge, which is however easy to be overlooked in traditional evaluation metrics. 
% For example, the model over-translates ``\textit{she spent one thousand}'' into ``\textit{她花了一千英镑}'' (English: \textit{she spent one thousand \textbf{pounds}}).
For example, the model mis-translates ``\textit{alas , he became sick from eating all of the peanut butter \textbf{on the ball}}'' into ``唉，他因为吃掉了球场上所有的花生酱而生病了'' (English: \textit{alas , he became sick from eating all of the peanut butter \textbf{on the field}}). 
With a minor mistake on the compound ``\textit{on the ball}'', the model achieves a sentence-level BLEU of 61.4, 
despite that the full sentence meaning is largely affected. 
% Despite the high BLEU score of this sentence, its meaning is totally incorrect.
In other words, the BLEU score of 69.58 can be misleading since novel compounds can be rare in the \textit{random test set}. 
Such mistakes in generalizing new compounds can severely hinder overall performance of translation engines in practice, as shown earlier in Table \ref{translation_engine}. 
% and faithfulness of the whole sentence translation, 
Also, we calculate BLEU for the original \textit{training} sentences that provide contexts for the CG test set (row 3).
The model achieves 99.74 BLEU, further demonstrating that the performance degradation is mainly caused by the unseen compounds.
\end{CJK*}

% For all patterns, the instance-wise error rate is 27.31\% and the compound-wise error rate is $61.62\%$, showing that a well-trained NMT model performs poorly in translating novel compounds.

Instance-wise, 27.31\% compounds are translated incorrectly. 
However, when aggregating all 5 contexts, $61.62\%$ compounds suffer at least one incorrect translation. 
% Even for those that are correctly inferred in all contexts, the correctness is not guaranteed in other contexts beyond the test set.  
% However, when looking at distinct compounds, $61.62\%$ suffers at least one error among 5 different contexts. 
This suggests that a well-trained NMT model is not robust in translating compounds, though all atoms within them are highly frequent in the training set. 
We also observe that the error rate of PP compounds, $37.72\%$, is much higher than the other two, $21.94\%$ and $22.25\%$, which we will discuss in detail in the following section. 
% The  are more likely to be mistranslated. 
% The error rate of PP patterns is much higher than the other two.
% The result also suggests that the model lacks the ability of compositional generalization, though compounds are merely syntactic constituents not more difficult sentences.
% all atoms within them are highly frequent in the training set. 

% Additionally, we also compute how many compounds are generalized successfully under all 5 contexts, i.e.,
% Additionally, we also compute aggregate error rate which reflects the ratio of compounds that are translated correctly in all 5 contexts, and the model fails to generalize as many as 67.41\% of all compounds. Moreover, even for compounds that are correctly inferred under 5 contexts, the model is likely to make mistakes in other contexts.

% The error rate of VP patterns is higher than that of NP while lower than PP. One possible reason is the different complexity of compounds in these three sets.

% Good performance in a random test set shows model's great generalization ability within the domain.  Whereas when there exists systematic differences between training and test sentences, the NMT model fails to generalize and obtains a much lower BLEU score.

\begin{table}
\centering
\small
% \setlength{\tabcolsep}{2.5pt}
% \begin{spacing}{1.0}
\begin{tabular}{c|c|c|c}
\hline
\multirow{2}{*}{\textbf{Test Set}} &  \multicolumn{2}{c|}{\textbf{Error Rate}} &  \multirow{2}{*}{\textbf{BLEU}} \\
\cline{2-3}
 & \textbf{Instance} & \textbf{Aggregate} & \\
\hline
\hline
Random-test & - & -  & 69.58  \\
\hline
Train & - & -  & 99.74  \\
\hline
CG-test &  27.31\% &  61.62\% & 48.66  \\
\hline
CG-test/NP &  21.94\% &  54.03\% & 51.29  \\
\hline
CG-test/VP &  22.25\% &  55.56\% & 47.55  \\
\hline
CG-test/PP &  37.72\% &  75.28\% & 47.14  \\
\hline
\end{tabular}
% \end{spacing}
\caption{
\label{tab_err_rate}
BLEU score and compound translation error rate on the \textit{random test} set and the \textit{CG test set}. 
}
\end{table}

\section{Analysis}
We conduct experiments to 
% analyze how the NMT model makes mistakes when translating novel compounds, 
explore in what situations the model is error-prone 
by considering compound frequency, compound length, compound structure, atom frequency, atom co-occurrence, 
and the complexity of external context.
% on compound translation. 

\subsection{Compound Frequency}
\begin{figure}[!t]
\centering
\includegraphics[width=0.85\linewidth]{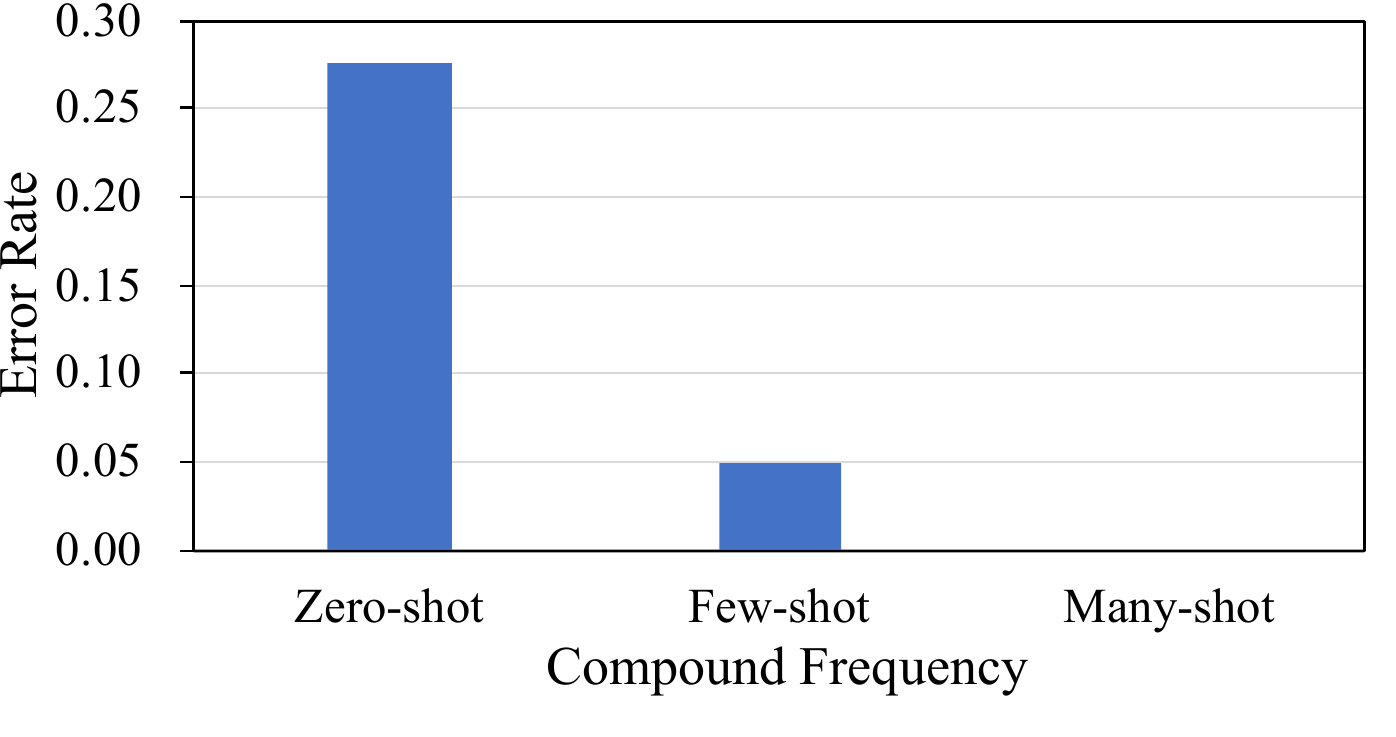}
\caption{
\label{compound-freq}
Effect of compound frequency on compound translation error rate.}
\end{figure}

Intuitively, compounds with higher frequencies in the training set are easier to infer. 
We classify compounds according to their frequency levels, including many-shots (frequency higher than 10), few-shots (frequency from 1 to 10) and zero-shot, 
and show the error rate for each bucket in Figure \ref{compound-freq}. 
% We observe that the model performs much better on the compounds that appear more than 10 times, obtaining a low error rate of 1.96\% and 3.64\% respectively. 
% We observe that the model makes no mistake translating compounds that appear more than 10 times in the training set. 
The model translates all the many-shots compounds correctly.
For few-shot compounds, translation error rate increases to $5.00\%$, but is still much lower than zero-shot compounds with an error rate of $27.53\%$. 
The result suggests the model is good at memorizing correspondence between sentence segments. However, the model deteriorates severely when test samples are unseen in the training set, which further confirms model's weakness in compositional generalization \cite{Lake:icml18}. 

% In other words, the NMT model can generalize very well for novel examples with a mixture of segments that have been seen during training. However, when examples are composed of more fine-grained atoms ????, the translation generalization ability deteriorates in a large degree.

\subsection{Compound Length}\label{comp_length}

\begin{figure}[!t]
\centering
\includegraphics[width=0.85\linewidth]{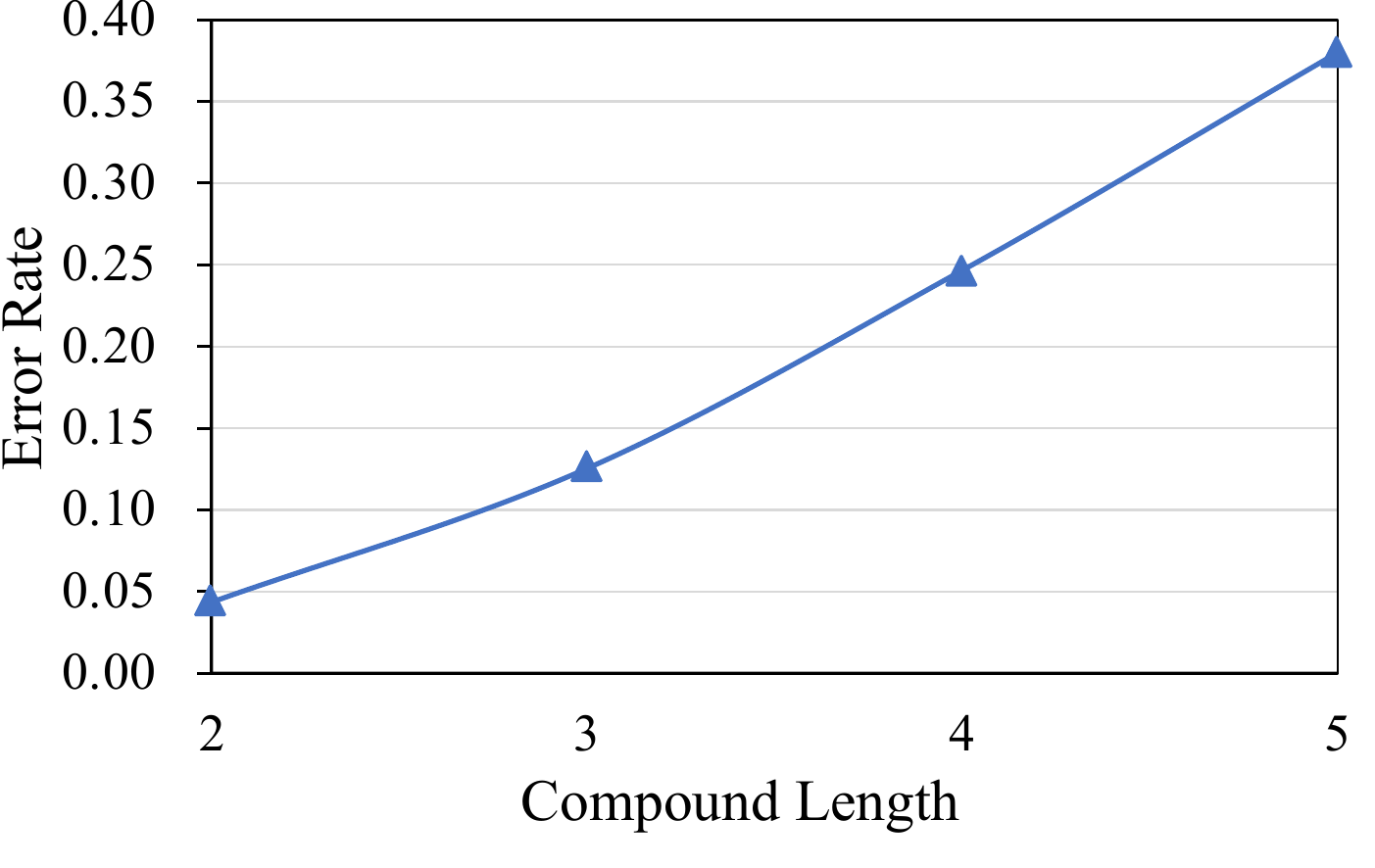}
\caption{
\label{compound-length}
Effect of compound length on compound translation error rate.}
\end{figure}

% We investigate the effects of compound length (the number of atoms in a compound), and observe that
As shown in Figure \ref{compound-length}, the error rate grows with the increase of compound length (i.e., the number of atoms in a compound). 
% the results are shown in 
Only 4.50\% of the shortest compounds are translated incorrectly, each of which consists of a determiner and a noun.
The error rate increases to 13.72\% when the compound length grows to 3 atoms (e.g., ``\textit{the smart lawyer}''). 
%  by adding an extra adjective or modifier,
The longest compounds contain a determiner, a noun, an adjective, a modifier and a preposition or verb in each of them, e.g., ``\textit{taking every special chair he liked}''. 
The error rate increases to 36.63\%, demonstrating that it is more difficult to generalize in longer compounds, 
which contain richer semantic information. 
% Note that {\it compound} in this paper is constituents, 
% Although we only examine compounds of a max length of 5, 
We conjecture that if the range of {\it compound} is further expanded, the error rate will be much higher.

\subsection{Atom Frequency}
\begin{figure}[!t]
\centering
\includegraphics[width=0.85\linewidth]{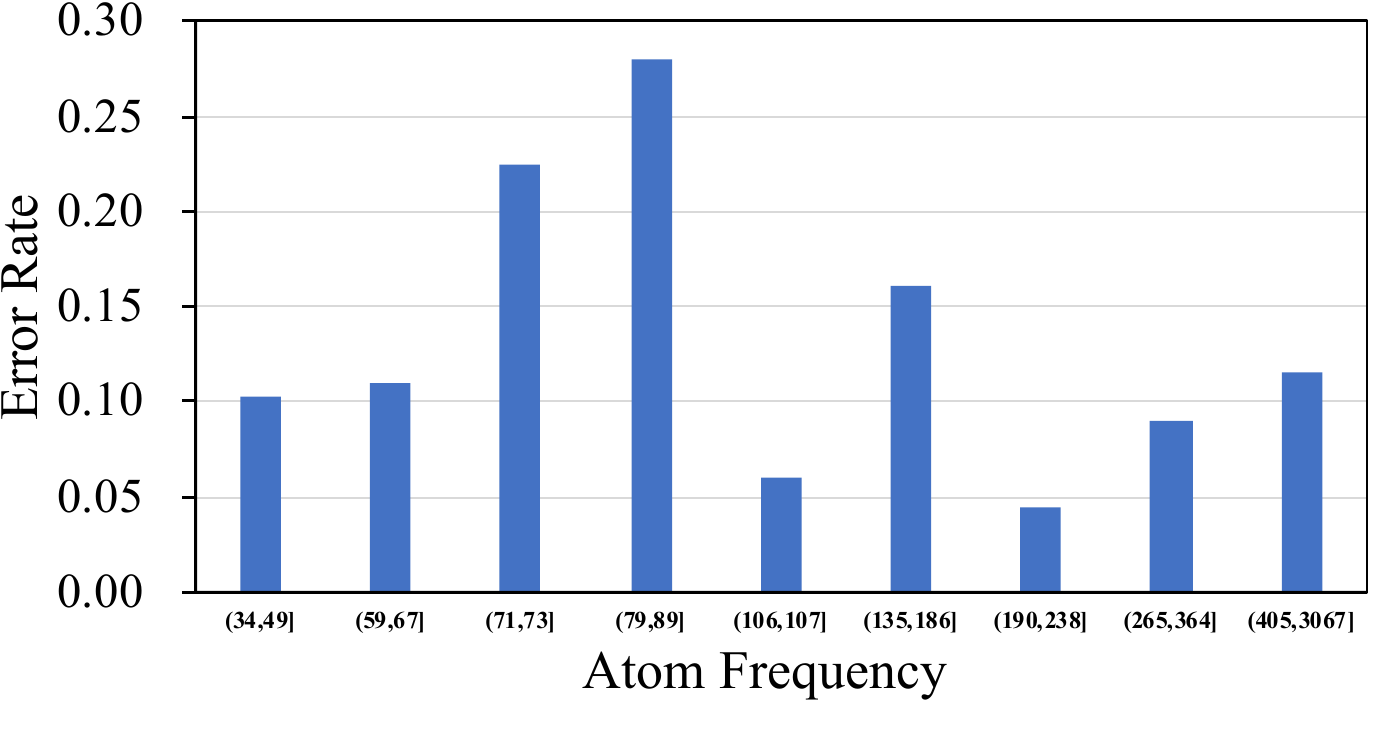}
\caption{
\label{atom-freq}
Effect of atom frequency on compound translation error rate.
}
\end{figure}
% Figure \ref{atom-freq} describes the effect of atom frequency on compound translation performance. 
We empirically divide compounds into multiple groups according to the minimum frequency of their atoms, where each group consists of similar numbers of compounds. 
The intuition is that the atom with low frequency might be difficult to translate and 
therefore hinders the whole compound translation. We fix the compound length to 3 in order to reduce effects of compound length. 

As shown in Figure \ref{atom-freq}, the error rate has no strong correlation with the atom frequency. 
This can be because all atoms in our corpus are simple and relatively frequent and thus it is easy for the NMT model to memorize the semantics of most atoms. 
Therefore, simply increasing atom frequency does not enhance model's generalization ability of novel compounds. 
We observe similar patterns for compounds of other lengths (Appendix \ref{Atom Frequency}).
% Hence, whether the new compounds can be translated correctly may depend on the difficulties of composition and context.

\subsection{Atom Co-occurrence}

\begin{figure}[!t]
\centering
\includegraphics[width=0.9\linewidth]{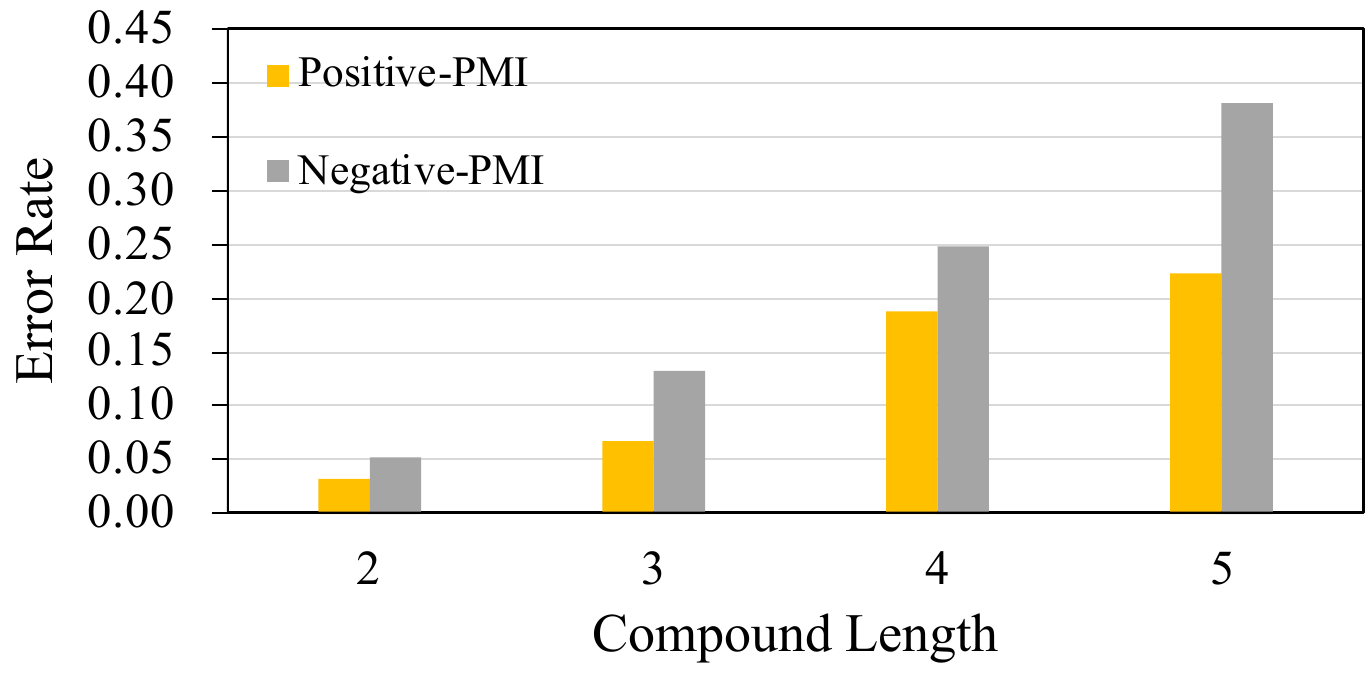}
\caption{
\label{occurrence}
Effect of atom co-occurrence on compound translation error rate.
}
\end{figure}

Although the NMT model may never see a compound, there can exist many local segments where atoms co-occur. 
For example, in the unseen compound ``\textit{the smart lawyer}'', ``\textit{smart}'' and ``\textit{lawyer}'' may occur within some training sentences.
Intuitively, the compounds of which atoms co-occur more frequently may be translated better.
We calculate pointwise mutual information (PMI) and compare error rates of compounds with positive or negative mean PMI scores (MPMI):
\begin{equation}
    {\rm MPMI}(C)=\frac{1}{M}\sum_{i=1}^{N-1}\sum_{j=i+1}^{N} {\rm PMI}(a_i,a_j),
\end{equation}
% \begin{align}
% \rm MPMI(C)&=\frac{1}{M}\sum_{i=1}^{N-1}\sum_{j=i+1}^{N} \rm PMI(a_i,a_j),\\
% \rm PMI(x,y)&=log\frac{p(a_i,a_j)}{p(a_i)p(a_j)},
% \end{align}
where $a_{i}$ is the $i$-th atom in the compound $C$, $N$ is the compound length, $M$ is the number of possible combinations of two atoms, 
% and PMI is calculated under 5-gram context (Appendix \ref{PMI score}).
and PMI score is computed as:
\begin{equation}
    PMI(x,y)=log\frac{p(a_i,a_j)}{p(a_i)p(a_j)},
\end{equation}
where the probabilities $p(a_{i})$ and $p(a_{i},a_{j})$ are obtained by dividing the number of n-grams in which one word or both words occur by the total number of n-grams\footnote{We use 5-gram here}.
% Next, $PMI(a_i,a_j)$ is derived by:
% \begin{equation}
% \end{equation}
% The probabilities $p(a_{i})$ and $p(a_{i},a_{j})$ are obtained by dividing the number of n-grams in which one word or both words occur by the total number of n-grams\footnote{We use 5-gram here}. 

We divide compounds into 4 groups by their length and compare error rates within each group. 
As shown in Figure \ref{occurrence}, across all groups, the error rates with positive mean PMI scores are lower than those with negative ones, verifying our hypotheses.

\begin{figure}[t]
\centering
\includegraphics[width=0.95\linewidth]{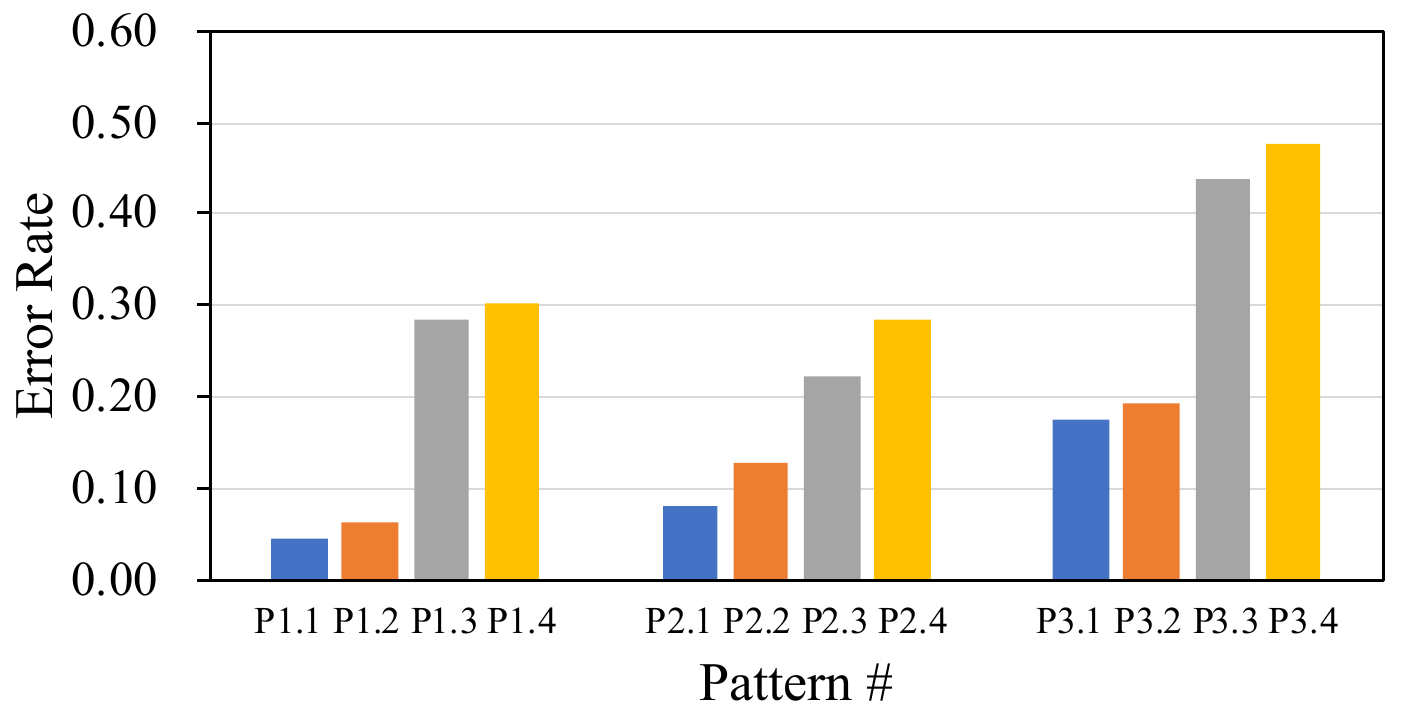}
\caption{
\label{compound-type}
Compound translation error rates of different patterns.}
\end{figure}

% \subsection{Compound Patterns}
\subsection{Linguistic Factors}
\begin{CJK*}{UTF8}{gbsn}
% As shown in table \ref{talbe_pattern}, the shortest and simplest compound contains a determiner and a noun, and such compound is embedded in all of the more complicated compounds. 
% As all compounds contain noun phrase, we divide compounds into 4 general groups based on their inner structures: noun, noun with adjective, noun with modifier and noun with both adjective and modifier.
% Each group is further split into 3 sub groups characterized by their syntactic roles in sentences: noun phrase, verb phrase and prepositional phrase. 
% In general, our compounds can be divided into NP compounds (Pattern 1.*), VP compounds (Pattern 2.*) and PP compounds (Pattern 3.*).
Figure \ref{compound-type} shows the error rates of all compound patterns in Table \ref{talbe_pattern}.
The MOD atom exerts salient influence on translation error rate. The error rate of compounds with MOD is 19.78\% higher than those without on average. 
In contrast, adding ADJ into compounds only increases error rate by 2.66\%. 
The major difficulty caused by MOD is word reordering. 
One can translate ``{\it the small dog}'' monotonically without adjusting word order. 
However, compounds like ``{\it the dog he liked}'' require the model to recognize ``{\it he liked}'' as MOD and put its translation before that of ``{\it the dog}'' in Chinese. 
We find many cases where the model translates such compounds without reordering or breaking the connection between nouns and modifiers.

% , i.e., ``他喜欢的''
%  into ``一只小狗''

Across these groups, we can see that the error rate of NP (Pattern 1.*) is generally lower than that of VP (Pattern 2.*) and PP (Pattern 3.*). 
Such phenomenon is more obvious for the patterns without MOD. 
The reason is that compounds in Pattern 1.* are generally shorter and contain less semantic and syntactic information. 
However, the error rates of Pattern 2.3 and 2.4 are lower than other patterns with MOD (i.e., Pattern 1.3, 1.4, 3.3 and 3.4), 
indicating the model performs better in ``\textit{V+DET(+ADJ)+NN+MOD}''.
This can be because under certain situations the MOD can be useful for correctly translating verbs, which are more commonly seen in the training set, e.g., ``\textit{\textbf{found} the chair \textbf{on the floor}}''.

We also observe that compounds of PP (Pattern 3.*) are more difficult to translate compared with VP (Pattern 2.*), although both types of compounds share the same compound length. 
In the training set, verbs typically have consistent translations, whereas the meanings of prepositions vary with contexts. Therefore prepositional compounds are more difficult to translate as more context information is required to ground their meanings.
\end{CJK*}
%  The meaning of a verb is more grounded than a preposition. less fluctuate than a preposition.
% verb, noun and context it connects with

\subsection{Effect of External Context}

\begin{figure}[!t]
\centering
\includegraphics[width=1.0\linewidth]{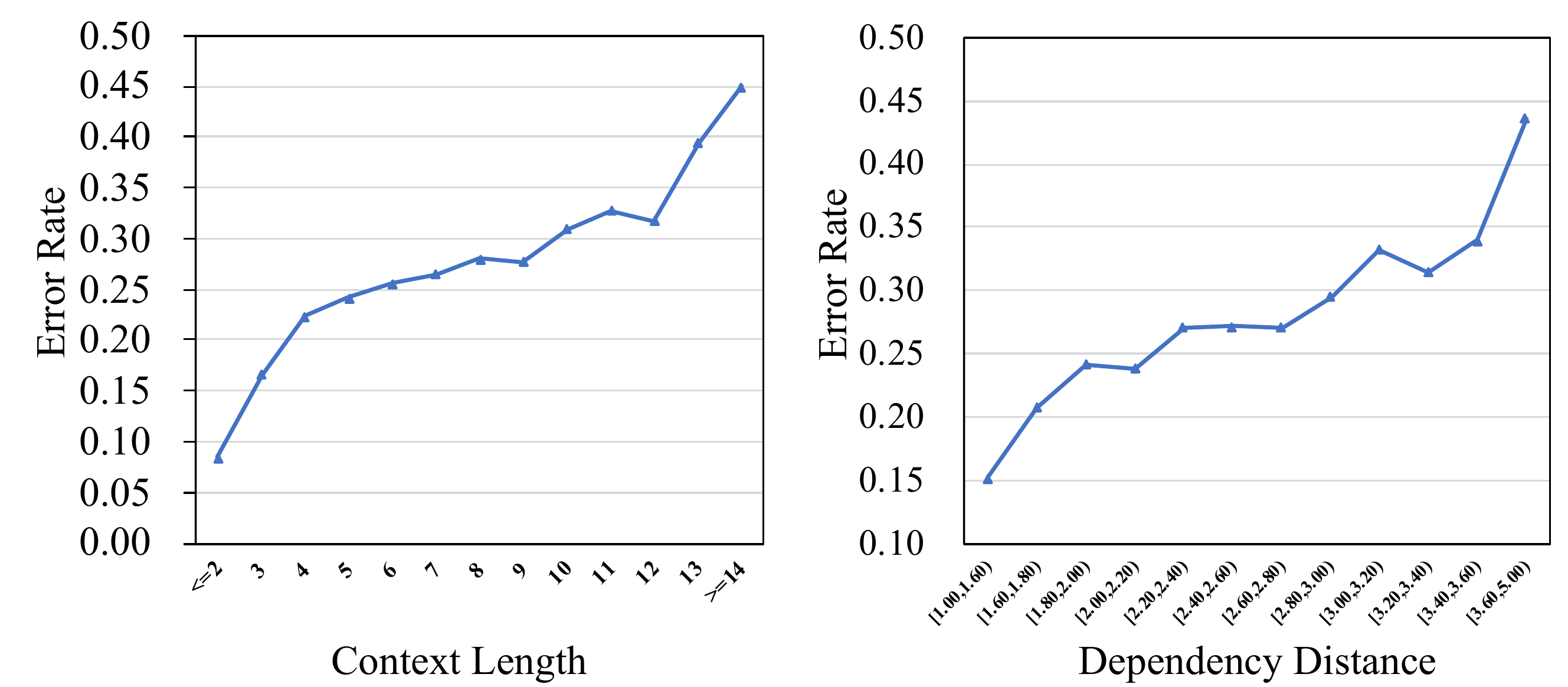}
\caption{
\label{context_comp}
Effect of external context on compound translation error rate.
}
\end{figure}

% is bound to a randomly sampled context and 
% beyond compound inner properties,
Due to the nature of NMT, the semantic representation of each compound is context-aware. 
Intuitively, translation of compounds is also influenced by external context, which is sentential in our case but can also be document-level in practice. 
% In our case, the external context is .
We investigate effects of context lengths and sentence comprehension difficulty. 
% We first experiment how compound translation error rate varies with respect to different context lengths. 
In particular, the context length is calculated by subtracting the sentence length by the number of words in the compound. Comprehension difficulty of the training sentences which provide contexts, is quantified by the dependency distance \cite{Haitao}:
% \begin{equation}
    ${\rm MMD}(x)=\frac{1}{N-1}\sum_{i}^{N}D_i$, 
% \end{equation}
where $N$ is the number of words in the sentence and $D_i$ is the dependency distance of the $i$-th syntactic link of the sentence. 
% We use Stanford Parser\footnote{https://github.com/stanfordnlp/stanza} to obtain dependency and calculate dependency distance for the sentences which generate context for compounds. 

The results are shown in Figure \ref{context_comp}. 
% There is a clear tendency that error rate increases as the context grows longer.
% As shown in Figure \ref{context_comp}, 
The translation error rate increases stably with the context length as well as the dependency distance. 
These observations demonstrate that the generalization for novel compounds correlates strongly with context complexity. 
Sentences with higher dependency distances are harder for model to comprehend during training. 
Given that our test sentences are restricted to 20 words, compositional generalization can be more challenging in practice where average sentence lengths can be much longer.

% Therefore, compounds in context from these complex sentences are generally more difficult to generalize. 

% The result suggests that the model fails to comprehend complex sentences during training, and therefore it is more difficult for it to translate novel compounds under such context.
% These observations demonstrate that the failure of compositional generalization occurs more in the complex sentence even though the model has seen the context in training. 

% \section{Case Study}

% \section{Evaluation}

\section{Conclusion}
We proposed a dedicated parallel dataset for measuring compositional generalization of NMT and quantitatively analyzed a Transformer-based NMT model manually. 
Results show that the model exhibits poor performance on novel compound translation, 
% This demonstrates that the problem of inability of compositionality still exists and is overlooked under transitional metrics.although the model is well trained and see all atoms frequently,
which demonstrates that the NMT model suffers from fragile compositionality, and it can be easily overlooked under transitional metrics.
To the best of our knowledge, we are the first one to propose a practical benchmark for compositionality of NMT, which can be a testbed for models tailored for this specific problem.

\section{Ethics Consideration}
As mentioned, we collected our data from Story Cloze Test and ROCStories Corpora that all are public to academic use, and they contain no sensitive information~\citep{mostafazadeh-etal-2016-corpus, mostafazadeh2017lsdsem}.
The legal advisor of our institute confirms that the sources of our data are freely accessible online without copyright constraint to academic use.
Our data construction involves manual annotation. 
Annotators were asked to post-edit machine translation and filter out samples that may cause ethic issues, which do not involve any personal sensitive information.

We hired 4 annotators who have degrees in English Linguistics or Applied Linguistics. Before formal annotation, annotators were asked to annotate $100$ samples randomly extracted from the dataset, and based on average annotation time we set a fair salary (i.e., 32 dollars per hour) for them. During their training annotation process, they were paid as well.

\section*{Acknowledgment}
Yue Zhang is the corresponding author.
We thank all reviewers for their insightful comments. This work is supported by National Natural Science Foundation of China (NSFC)
under grant No.61976180 and a grant from Lan-bridge Information Technology Co., Ltd. 
We thank colleagues from Lan-bridge for examining data and evaluating results. 
Major contributors include Xianchao Zhu, Guohui Chen, Jing Yang, Jing Li, Feng Chen, Jun Deng and Jiaxiang Xiang.
% calculated our average annotation time so we could set a fair salary for annotators’ training annotation. During their training annotation process, they were paid as well. And we also calculated the average annotation time for each dialogue during their training, based on which we determined the final average salary of annotation was around 9.5 dollars per hour. This hourly salary was the same for manual checking. All of our annotators took this annotation as a part-time job.
% We intend this contribution to provide a standard benchmark for robustness to noise in MT and foster research on models, dataset and evaluation metrics 

% In the future, we will attempt to explore novel NMT architectures being compositionality.
% We intend this contribution to provide a dedicated benchmark for compositionality in NMT.
% Although NMT models have achieved competitive performance evaluated by BLEU scores, they still lack the ability of compositional generalizaiton. 

% 严重的，存在的，被忽略的

% 我是第一个xx

% 我们的data，models都得来测一下。

% \section*{Acknowledgments}
% The acknowledgments should go immediately before the references. Do not number the acknowledgments section.
% \textbf{Do not include this section when submitting your paper for review.}

\bibliographystyle{acl_natbib}
\bibliography{acl2021,anthology}
% \newpage
\clearpage
\appendix

\section{Atom Frequency}\label{Atom Frequency}
\begin{figure}[H]
\centering
\includegraphics[width=0.8\linewidth]{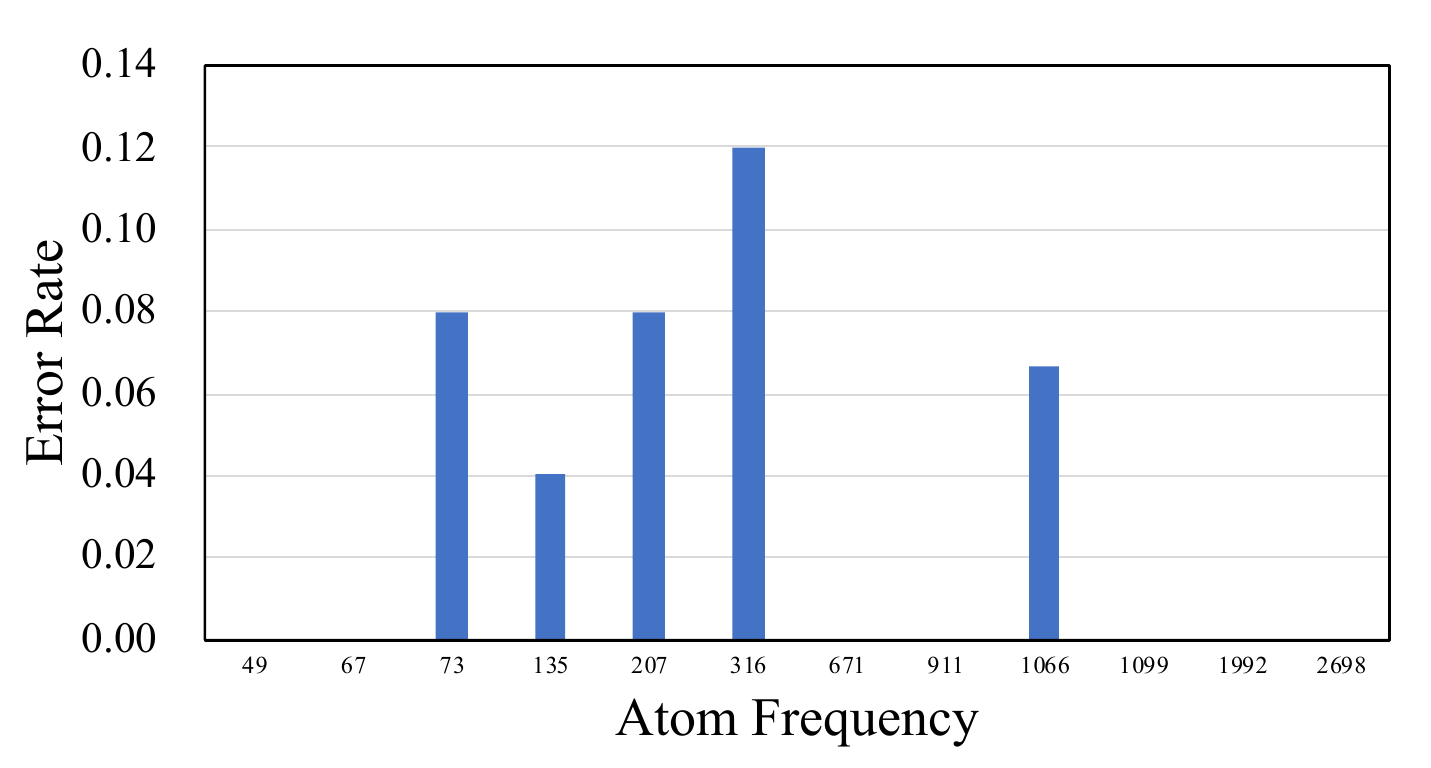}
\caption{
\label{atom-freq2}
Effect of atom frequency with compound length fixed to 2.
}
\end{figure}

\begin{figure}[H]
\centering
\includegraphics[width=0.8\linewidth]{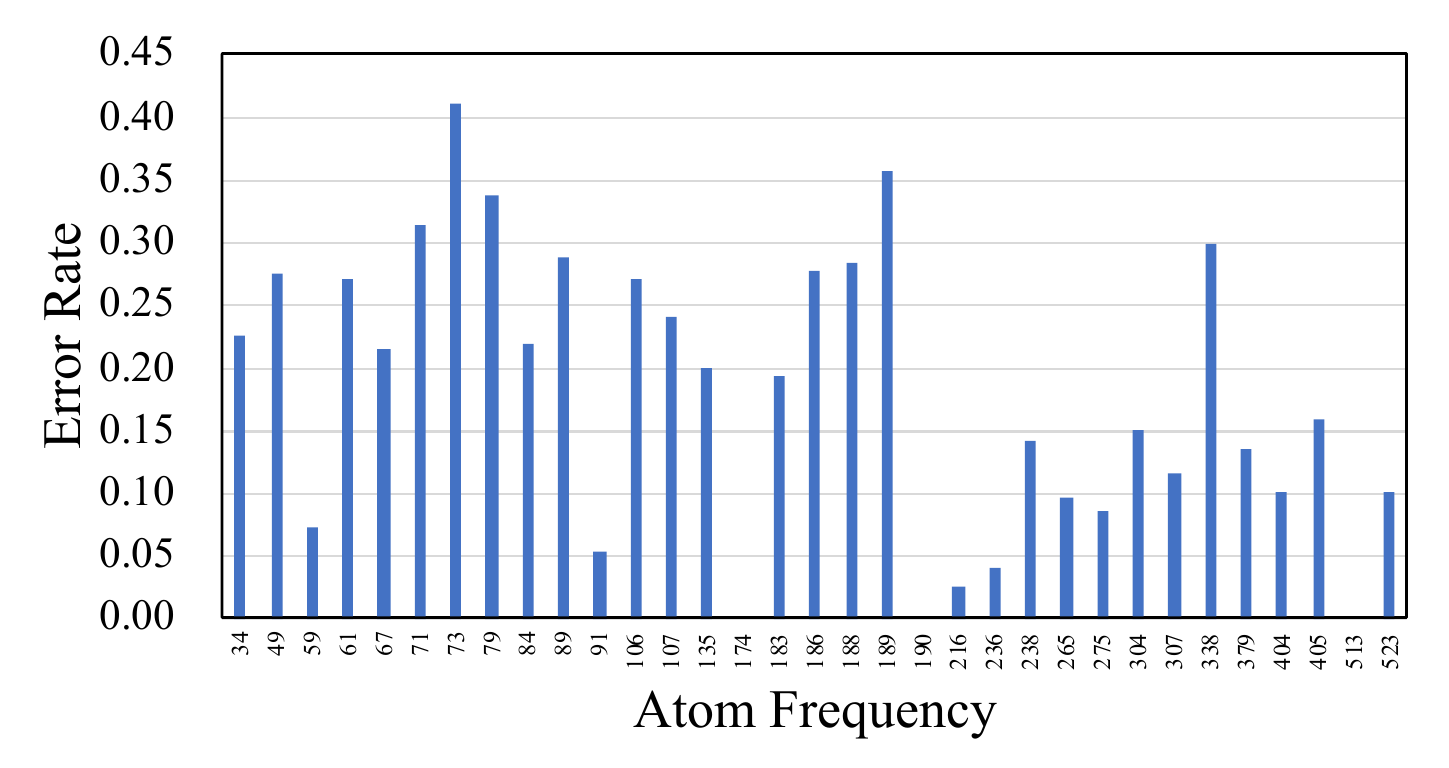}
\caption{
\label{atom-freq4}
Effect of atom frequency with compound length fixed to 4.
}
\end{figure}

\begin{figure}[H]
\centering
\includegraphics[width=0.8\linewidth]{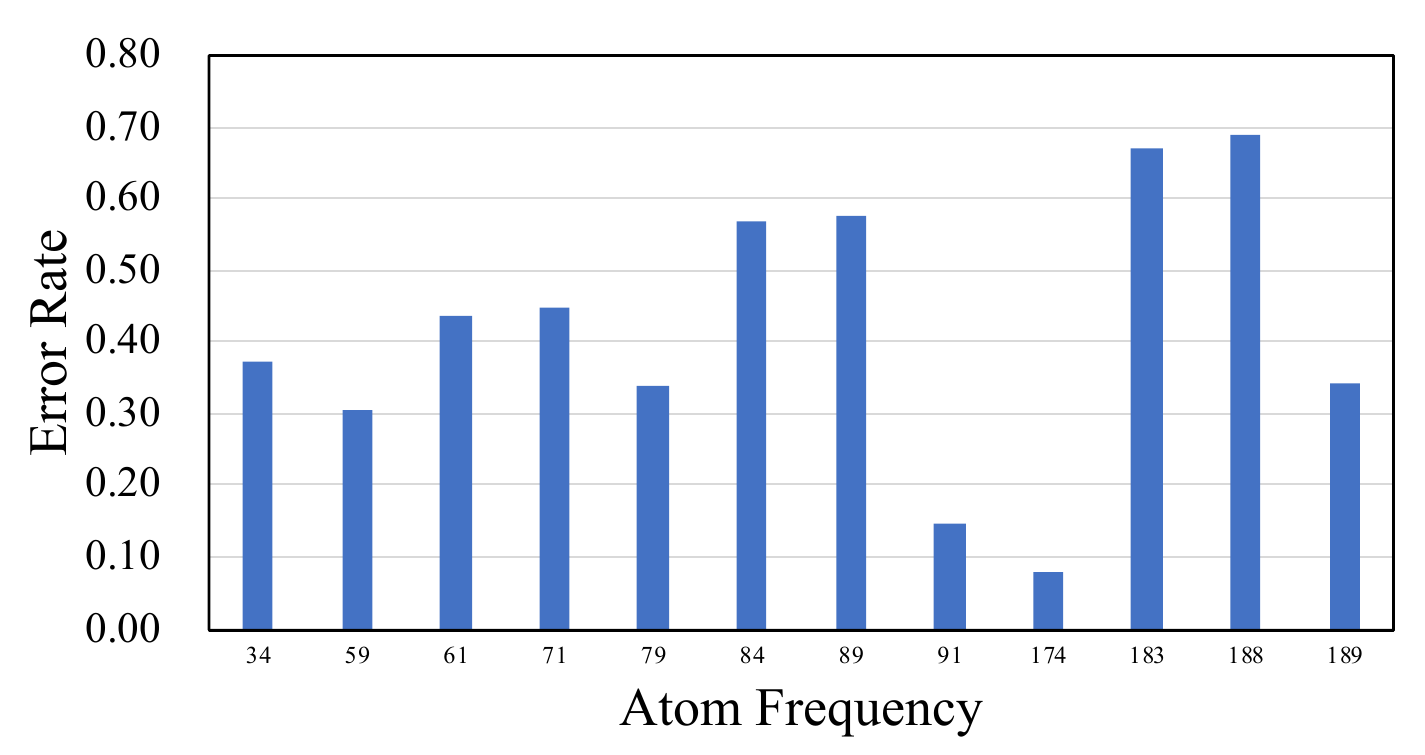}
\caption{
\label{atom-freq5}
Effect of atom frequency with compound length fixed to 5.
}
\end{figure}

For compounds of other lengths, we also compute their error rates with respect to minimum atom frequency. As shown in Figure \ref{atom-freq2}, \ref{atom-freq4} and \ref{atom-freq5}, the error rate does not correlates with atom frequency across all compound lengths.

\section{Data Statistics}\label{Data Statistics}

Table \ref{tab_data_statistics_1} and Table \ref{tab_data_statistics_2} lists statistics of several monolingual data sources, compared with the data source (ROC-Filter) used in constructing the CoGnition dataset. We can see that our dataset has both shorter sentences and vocabulary made up of more frequent words.

\begin{table}[H]
\centering
\small
\setlength{\tabcolsep}{2.5pt}
% \begin{spacing}{1.0}
\begin{tabular}{l|c|c|c}
\hline
\diagbox{\textbf{Data}}{\textbf{Property}} & \textbf{Vocab} & \#\textbf{Tokens} & \#\textbf{Sents}  \\
\hline
\hline
WMT17 En-Zh & 1,201,752 & 518,286,577 & 20,616,495  \\
\hline
IWSLT17 En-Zh & 70,950 & 4,715,201 & 231,266  \\
\hline
ROC-Original & 42,458 & 5,283,521 & 532,093  \\
\hline
ROC-Filter & 2,000 & 2,096,524 & 216,246 \\
\hline
\end{tabular}
% \end{spacing}
\caption{
\label{tab_data_statistics_1}
Statistics of data sources: vocabulary size, number of tokens and number of sentences. 
}
\end{table}

\begin{table}[H]
\centering
\small
\setlength{\tabcolsep}{2.5pt}
% \begin{spacing}{1.0}
\begin{tabular}{l|c|c|c}
\hline
\diagbox{\textbf{Data}}{\textbf{Property}} & \textbf{Avg Len} & \textbf{Avg Freq} & \textbf{Min Freq} \\
\hline
\hline
 WMT17 En-Zh & 25.1 & 431.3 & 1 \\
\hline
 IWSLT17 En-Zh & 20.4 & 66.5 & 1 \\
\hline
 ROC-Original & 9.3 & 124.4 & 1 \\
\hline
 ROC-Filter & 9.7 & 1048.3 & 35 \\
\hline
\end{tabular}
% \end{spacing}
\caption{
\label{tab_data_statistics_2}
Statistics of data sources: average sentence length, average token frequency and minimum token frequency. 
}
\end{table}

\section{Lexicon}\label{Lexicon}
Part of the lexicon for automatic evaluation is shown in Table \ref{talbe_dict}.
\begin{CJK*}{UTF8}{gbsn}
\begin{table}[H]
\centering
\small
% \setlength{\tabcolsep}{2.5pt}
% \begin{spacing}{1.0}
\begin{tabular}{l|c}
\hline
Atom & Lexical Translation \\
\hline
dog	& 狗/犬 \\
\hline
doctor & 医生 \\
\hline
sandwich & 三明治 \\
\hline
hat & 帽 \\
\hline
waiter & 服务员 \\
\hline
lawyer & 律师 \\
\hline
peanut & 花生 \\
\hline
farmer & 农夫/农场主/农贸市场/农民 \\
\hline
small & 小 \\
\hline
red & 红 \\
\hline
dirty & 脏 \\
\hline
lazy & 懒 \\
\hline
smart & 聪明/明智/智能 \\
\hline
the & - \\
\hline
every & 每/所有 \\
\hline
any & 任何 \\
\hline
another & 另/又/再/还/别 \\
\hline
each & 每 \\
\hline
he liked & 他喜欢的 \\
\hline
at the store & 店里/商店 \\
\hline
\end{tabular}
% \end{spacing}
\caption{
\label{talbe_dict}
Lexicon for automatic evaluation.
}
\end{table}
\end{CJK*}

\end{document}